\documentclass{article}
\usepackage{ijcai24}
\usepackage{times}
\usepackage{soul}
\usepackage{url}
\usepackage[hidelinks]{hyperref}
\usepackage[utf8]{inputenc}
\usepackage[small]{caption}
\usepackage{graphicx}
\usepackage{amsmath}
\usepackage{amsthm}
\usepackage{booktabs}
\usepackage{algorithm}
\usepackage{algorithmic}
\usepackage[switch]{lineno}
\usepackage{placeins}

\title{Examining Monitoring System: Detecting Abnormal Behavior \\In Online Examinations}

\author{
    Dinh An Ngo$^{1}$\thanks{These authors contributed equally to this work}
    \and
    Thanh Dat Nguyen$^{1}$\footnotemark[1]
    \and
    Thi Le Chi Dang$^{1}$\footnotemark[1]
    \and Huy Hoan Le$^{1}$\footnotemark[1]
    \and Ton Bao Ho$^{1}$\footnotemark[1]
    \and \\Vo Thanh Khang Nguyen$^{2}$
    \and Truong Thanh Hung Nguyen$^{2,3}$
    \affiliations
    $^1$FPT University, Quy Nhon, Vietnam\\
    $^2$Quy Nhon AI, FPT Software, Vietnam\\
    $^3$Analytics Everywhere Lab, University of New Brunswick, Canada
    \emails
    \{AnNDQE170066, DatNTQE170110, ChiDLTQE170156, HoanLHQE170006,\\ BaoHTQE170007\}@fpt.edu.com,
    \{KhangNVT1, HungNTT\}@fpt.com
}

\begin{document}

\maketitle

\begin{abstract}
Cheating in online exams has become a prevalent issue over the past decade, especially during the COVID-19 pandemic. To address this issue of academic dishonesty, our ``Exam Monitoring System: Detecting Abnormal Behavior in Online Examinations'' is designed to assist proctors in identifying unusual student behavior. Our system demonstrates high accuracy and speed in detecting cheating in real-time scenarios, providing valuable information, and aiding proctors in decision-making. This article outlines our methodology and the effectiveness of our system in mitigating the widespread problem of cheating in online exams.
\end{abstract}

\section{Introduction}
The increase in online exams, especially during the COVID-19 pandemic~\cite{newton2023common}, has led to a significant issue: widespread cheating in online exams~\cite{dendir2020cheating,bucciol2020cheating}. Candidates take advantage of the virtual setting by frequently scanning their surroundings, searching for external documents, and copying from peers. As a result, manual supervision becomes challenging in the crowded online exam environment. To tackle this problem, we developed an “Exam Monitoring System: Detecting Abnormal Behavior in Online Examinations” that ensures peripheral conditions to identify abnormal behavior, thereby ensuring fairness in online exams. The current system is applied to proctored online exam rooms to address limitations in existing proctoring methods, particularly when dealing with a growing number of candidates, thus reducing proctor effort. Our contributions include:
\begin{itemize} 
\item Enhancing fairness and effectiveness in online exam monitoring. 
\item Managing exam cheating using AI-driven approaches. 
\item Facilitating informed decision-making with AI assistance. \end{itemize}

\section{Related Work}

In the field of online examination systems, various studies have been conducted to detect abnormal behaviors during the examination process \cite{abbas2022systematic,mohammed2022eproctoring,Muzaffar6310,moukhliss2023intelligent}. An early attempt~\cite{hendryli2016classifying} proposes a method using a supervised dynamic hierarchical Bayesian model called MCMCLDA, which models the activities as sequences of motion patterns based on interest points detected by MODEC or Harris3D, achieving an accuracy of 57\%. \cite{kuin2018} analyzed electronic test videos using VGG16~\cite{simonyan2014very}, Inception-v4~\cite{szegedy2017inception}, and MobileNets~\cite{howard2017mobilenets}. They achieved 96.8\% accuracy in traditional training, but this dropped to 67.1\% in real-time testing. \cite{asker} developed a multi-model online test supervision system, which includes a facial recognition model with 90\% accuracy, YOLOv3 for mobile detection with 81\% accuracy, and pupil tracking with 92.5\% training accuracy, which drops to over 75\% during inference. 

Our system stands out by achieving an accuracy of 78.5\% on a distinctive test set comprising different individuals from the datasets, demonstrating its performance and applicability in real-world scenarios. A key feature of our system is its optimization for real-time performance, achieving a speed of 27 frames per second (FPS). This addresses the critical need for fast and efficient monitoring in online examinations. By prioritizing real-time processing speed, our system offers proctor support solutions, reducing the effort required by proctors during exams and promoting fair examination practices.

\section{System}
\begin{figure}[h]
    \centering 
    \includegraphics[width=0.45\textwidth]{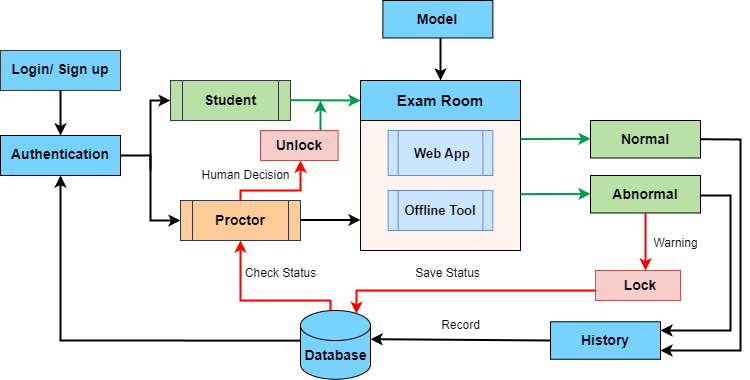} 
    \caption{System Overview: Before entering the exam room, individuals authenticate their identity as a student or proctor. The system aids proctors by monitoring student behavior, issuing warnings, and capturing images of rule breaches. If a student breaks the rules more than three times, the quiz locks. The proctor then decides to either unlock or end the exam.} 
    \label{sec:system_overview} 
\end{figure}

Our proposed system offers a solution to address fraud challenges in the digital environment. This section provides a concise overview of the system's functions and components, offering a comprehensive understanding of the proposed framework in Figure \ref{sec:system_overview}.

% \subsection{Demonstrations}
% The proposed system caters to both students and proctors.

% \begin{itemize}
% \item\textbf{Students' Role:} Students access the exam quiz, install a behavior-detection extension and take the exam. Warnings may be issued for abnormal behavior.

% \item\textbf{Proctors' Role:} Proctors can create exam rooms, invite students to the rooms, and upload questions. During the exam, warnings are issued for abnormal behavior. If a candidate consistently violates the exam rules, their test will be blocked. The proctor will review the candidate's actions and unlock the test if there has been an unintentional violation. The dashboard provides summary visuals.
% \end{itemize}

\subsection{Functionalities}
\subsubsection{Abnormal Behavior Detection Tool}
\begin{figure}[h]
    \centering 
    \includegraphics[width=0.45\textwidth]{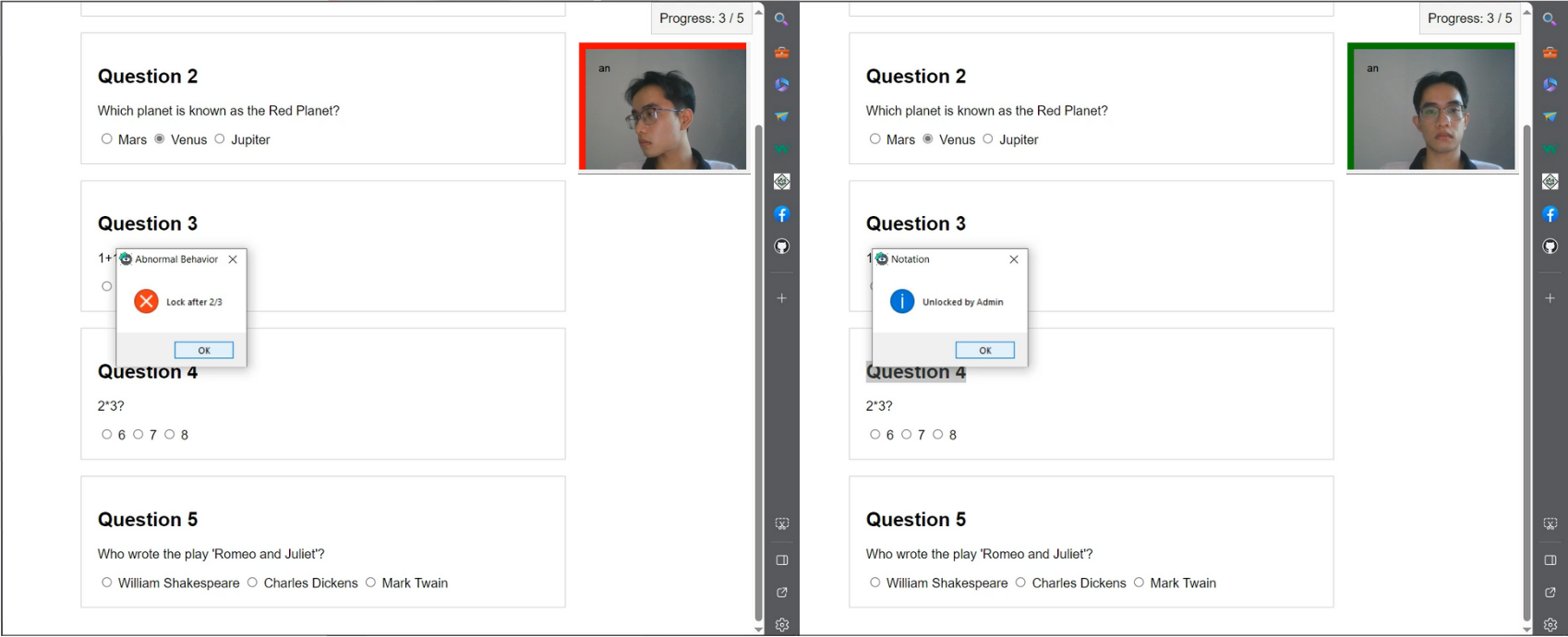} 
    \caption{The student’s screenshot during the exam is displayed. The left side shows a photo of the violation, with the system issuing a warning to the students. The right side shows a photo taken when the proctor resumes the exam after a suspension due to the student committing more than three violations.} 
    \label{sec:deploy_demo} 
\end{figure}

The Exam Room function on the website serves as a secure space for online exams, enhanced by the Abnormal Behavior Detection Tool we deployed as a pilot for a class with 20 students, demonstrated in Figure \ref{sec:deploy_demo}.
% , and plan to deploy widely in April 2024. 
Our tool includes the following functions:

\begin{itemize}
\item \textbf{Real-time Monitoring:} Our system employs a computer vision model (Section~\ref{sec:model}) to monitor students’ webcam feeds during exams and classify behaviors as normal or abnormal. Proctors can review these flagged behaviors and may contact students as necessary.
\item \textbf{Abnormality Alerts:} The system generates immediate alerts for abnormal behaviors, which are displayed on an admin dashboard along with the captured images for proctor review. If abnormal behavior surpasses a certain threshold, proctors receive notifications on their dashboard and can lock the exam, preventing further attempts until they intervene.
\end{itemize}

% \subsection{Model}\label{sec:model}
% Our model pipeline runs as follows (Figure \ref{sec:model_pipeline}):

% \begin{itemize}
% \item \textbf{Dataset Development:}
% We manually created a unique dataset because of the lack of public datasets for detecting behavior in online exams. This dataset is used for training and evaluating the model of the proposed system.
 
% \item \textbf{Data Collection:}
% Adopting a meticulous approach, the dataset comprises 1200 images collected using a laptop webcam with default settings. This dataset captures several cheating scenarios, such as head movements, eye movements, and individuals wearing masks or sunglasses in different locations and lighting conditions, ranging from classrooms to coffee shops, to enhance the realism of the dataset. Figure \ref{fig:datasetsamples} provides visual details about some sample data in the dataset.

\subsection{Model Development}\label{sec:model} 
\begin{figure*}[h]
    \centering 
    \includegraphics[width=0.8\textwidth]{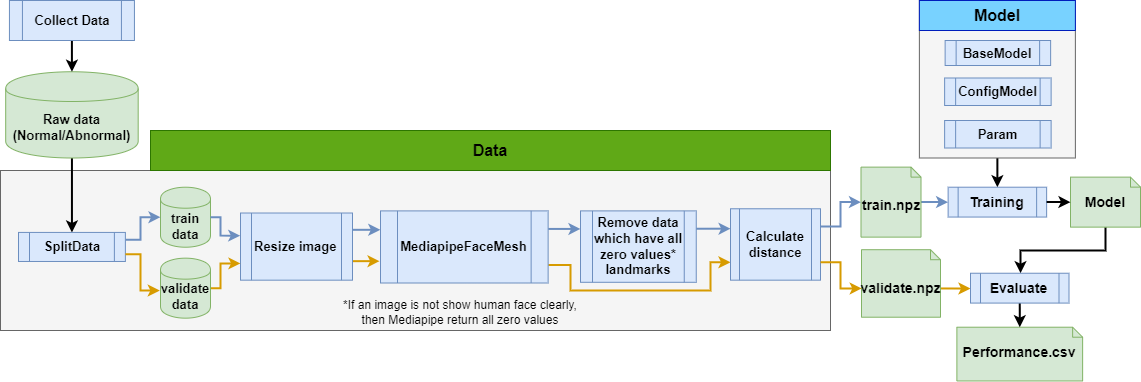} \caption{The pipeline follows a sequence of steps from data collection to model training. Data is collected from the laptop camera, split, and resized. Facial landmarks are extracted using Mediapipe for initial preprocessing. Images with all zero values are removed from the training set. The Euclidean distance formula is used to calculate distances between 19 selected landmark points. The final dataset is then used for training and validation.} 
    \label{sec:model_pipeline} 
\end{figure*}

In this section, we detail the model pipeline (as illustrated in Figure \ref{sec:model_pipeline}) and discuss the steps involved in its creation and operation:

\subsubsection{Dataset Development} Due to the absence of public datasets for detecting abnormal behavior in online exams, we manually curated a unique dataset. This dataset serves as the foundation for training and evaluating our proposed system’s model.

\subsubsection{Data Collection} 
\begin{figure}[h]
    \centering
    \includegraphics[width=0.35\textwidth]{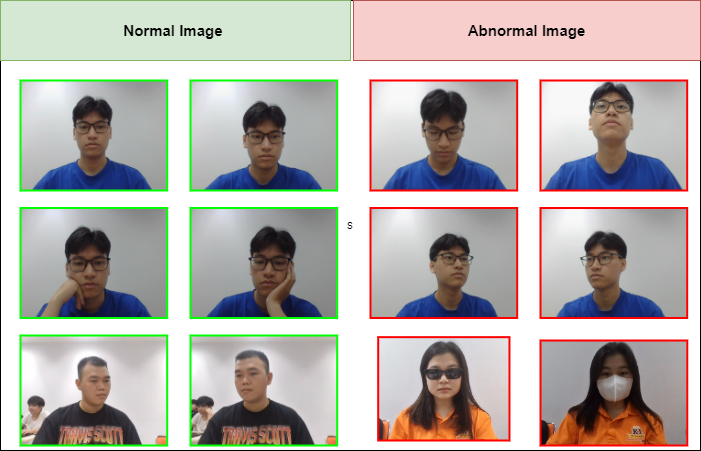}
    \captionof{figure}{Sample data of Normal (left) and Abnormal (right) class.}
    \label{fig:datasetsamples}
\end{figure}

We collected 1200 images using a laptop webcam with default settings, adopting a comprehensive approach. The dataset encapsulates various cheating scenarios, such as head and eye movements, and individuals wearing masks or sunglasses. These images were captured in diverse locations and lighting conditions, including classrooms and coffee shops, to enhance the dataset’s realism. Visual details of some sample data in the dataset are provided in Figure \ref{fig:datasetsamples}.

\subsubsection{Data Preprocessing}
The raw data was preprocessed and divided into training and validation sets at an $80:20$ ratio. The images were cropped to a resolution of $640 \times 480$ to focus on the user, excluding unnecessary elements like walls or outdoor scenery. Mediapipe~\cite{lugaresi2019mediapipe} was used to extract facial landmark points. Images where Mediapipe failed to recognize faces, yielding all zero landmark points, were removed from the dataset. After extensive experimentation, we identified 17 facial points and 2 eye center points. The Euclidean distance formula was used to calculate pairwise distances between these points, generating 171 key features for model training.

\subsubsection{Model Experiment} 
\begin{figure}
    \centering
    \includegraphics[width=0.45\textwidth]{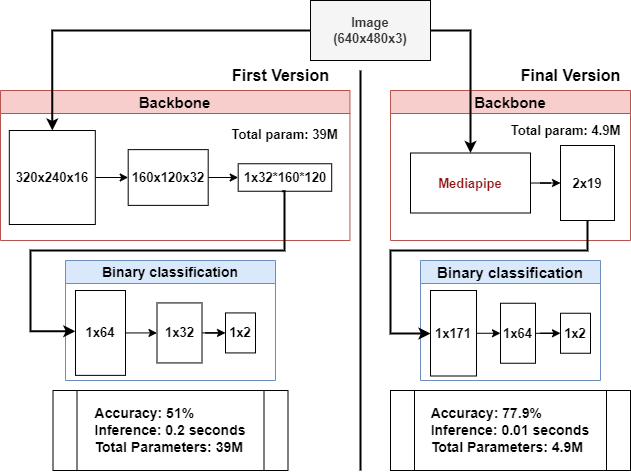}
    \caption{Model architecture comparison between CNN and Mediapipe. We use Mediapipe to extract features instead of letting the model autonomously learn them, facilitating the learning process and reducing external environmental influences.}
    \label{fig:model_architecture}
\end{figure}

We initially used a Convolutional Neural Network (CNN)~\cite{schmidhuber2015deep} for feature extraction and binary classification, but the results were not satisfactory. We then tried VGG16, which was slow, and MobileNetv3-Small~\cite{howard2019searching}, which was faster but lacked accuracy. Ultimately, we adopted a model that combines Mediapipe Face Mesh~\cite{kartynnik2019real} for facial landmark extraction and a fully connected (FC) layer for binary classification. The architecture details are depicted in Figure \ref{fig:model_architecture}. We also tested various feature selection methods with Mediapipe to optimize efficiency, as detailed in the Data Preprocessing Section. The results are presented in Table \ref{tab:feature_selection}).

\subsubsection{Model Evaluation} 
The model’s performance was assessed using several metrics on the test dataset after 100 epochs of training. We prioritized a simple model with a fast inference speed for real-time implementation. The real-time test performance is evaluated in Section \ref{sec:results}.

% The pipeline illustrates the sequential steps from data collection to the training model. The data collected from the laptop camera will be split. Before entering the training phase, we will resize the images and extract key points. The distance between these keypoints will be utilized by the Mediapipe for initial preprocessing before further input into the model. Rather than allowing the model to autonomously learn the feature class, we opted to use Mediapipe to extract initial features. This approach facilitates the model's learning process and minimizes the influence of the external environment.

\section{Results}\label{sec:results}
This section showcases our study's results, using benchmarks to evaluate the proposed system's performance and efficacy.

We selected the Mediapipe model and FC architecture for our system to prioritize speed. This combination yields an impressive inference time of just 0.01 seconds, facilitating rapid image processing. With this configuration, we attain an average processing speed of 25 to 27 FPS on 20 clients using a virtual Linux machine with 8 MB of RAM and 1 GB of ROM. The server specification includes an Intel Core i5-10400F CPU and 8 GB of RAM, ensuring real-time detection capability on most student devices.

By utilizing the Mediapipe and FC layer with appropriate features, our system proves its effectiveness by achieving an accuracy of 78.5\% on the test set, which includes a distinct person from the datasets and comprises 200 frames. We successfully make 157 correct predictions.

\begin{table}[h]
\centering
\resizebox{0.9\linewidth}{!}{
\begin{tabular}{ccccc} 
    \toprule
    Model selection & Accuracy↑ & Parameters↓ & Inference time↓ & FPS↑\\ 
    \midrule
    CNN backbone + FC & 51\% & 39M & 0.2s & 14 \\ 
    
    VGG16 & \textbf{79\%} & 138M & 0.34s &  5 \\ 
 
    MobileNetv3-Small & 68\% & \textbf{2.5M} & \underline{0.03s} & \underline{26}\\ 
 
    Mediapipe + FC & \underline{77.9}\% & \underline{4.9M} & \textbf{0.01s} & \textbf{27} \\ 
    \bottomrule
\end{tabular}
}
\caption{Result in different models on the validation set. For each metric, bold letters represent the best outcome, and underlined letters represent the second best. The arrows $\uparrow/\downarrow$ indicate higher or lower scores are better.}
\label{tab:feature_selection}
\end{table}

\begin{table}[h]\label{sec:Table2}
\centering
\resizebox{0.75\linewidth}{!}{
    \begin{tabular}{cccc} 
        \toprule
        Feature selection & Accuracy & Precision & Recall \\ 
        \midrule
        \begin{tabular}[c]
        {@{}c@{}}478 keypoints \\(2*478 dimensions)\end{tabular} & 61\% & 0.7568 & 0.7482 \\
        \midrule
        \begin{tabular}[c]
        {@{}c@{}}19 keypoints \\ (2*19 dimensions)\end{tabular} & 65\% & 0.6555 & 0.6547 \\
        \midrule
        \begin{tabular}[c]
        {@{}c@{}}19 pair-wise distances \\ (1*171 dimensions)\end{tabular} & \textbf{77.9\%} & \textbf{0.8143} & \textbf{0.7651} \\
        \bottomrule
    \end{tabular}
}
\caption{\label{tab:feature_selection}Result in different feature selection methods with Mediapipe + FC model on the validation set. For each metric, bold letters represent the best outcome.}
\end{table}

\section{Conclusion And Future Work}

Our ``Examining Monitoring System: Detecting Abnormal Behavior In Online Examinations'' represents a significant step forward in enhancing exam integrity and addressing abnormal behavior in online exams. Leveraging advanced vision AI models, our project aims to assist proctors in detecting abnormal candidate behaviors. In future work, we plan to improve the system to better serve the diverse needs of individuals and businesses. This improvement will include incorporating multi-label classification, using facial recognition to verify candidates, and object recognition to identify abnormal behaviors. The ultimate goal is to create an exam environment without the presence of a proctor.

\newpage
\bibliographystyle{named}
\bibliography{ijcai24}
% \bibliographystyle{unsrt}
% \cite{alkhalisy2023}
% \cite{atoum2017}
% \cite{ozgen2021}
% \cite{hu2018}
% \cite{ibrahim2018}
% \cite{oviya2022}
% \cite{alrawi2022}
% \cite{bartlett2002}
% \cite{rowley1998}
% \cite{borges2013}
% \cite{yang2022}
% \cite{samal1992}
% \cite{krekovic2012}
% \cite{breitenstein2008}
% \cite{mathapati2017}
% \cite{ching2021}
% \cite{bobick2002}

\end{document}